%% file: main.tex
\documentclass[letterpaper, 10pt,conference,twocolumn]{ieeeconf}

\input{packages}
\IEEEoverridecommandlockouts  
\input{newcommands}

\newcommand{\POMDP}{\mathcal{P}}
\newcommand{\stateSpace}{\mathcal{S}}
\newcommand{\observationSpace}{\mathcal{O}}
\newcommand{\actionSpace}{\mathcal{A}}
\newcommand{\transitionModel}{\mathcal{T}}
\newcommand{\observationModel}{\Omega}
\newcommand{\rewardFunction}{\mathcal{R}}
\newcommand{\externalTau}{\tau_\text{ext}}
\newcommand{\belief}{b}
\newcommand{\goalSet}{\mathcal{G}}
\newcommand{\displacement}{e}
\newcommand{\displacementSet}{\mathcal{D}}
\newcommand{\rotationMatrix}{\text{Rot}}
\newcommand{\feedforwardTau}{\tau_{ff}}

\title{POMDP-Guided Active Force-Based Search for Robotic Insertion}
\author{Chen Wang$^{1, 2}$, Haoxiang Luo$^{1}$, Kun Zhang$^{3}$, Hua Chen$^{1}$, Jia Pan$^{2}$, Wei Zhang$^{1}$
\thanks{$^1$School of System Design and Intelligent Manufacturing, Southern University of Science and Technology, Shenzhen, China. Emails: {\tt luohxisme@gmail.com, chenh6@sustech.edu.cn, zhangw3@sustech.edu.cn}}
\thanks{$^2$Department of Computer Science, the University of Hong Kong, Hong Kong. Emails: {\tt cwang5@cs.hku.hk, jpan@cs.hku.hk}}
\thanks{$^3$Cheng Kar-Shun Robotics Institute, Hong Kong University of Science and Technology, Clear Water Bay, Hong Kong. Email: {\tt kun.zhang@connect.ust.hk}}
\thanks{Corresponding author: Wei Zhang and Jia Pan}
\thanks{This work was supported in part by National Natural Science Foundation of China under Grant No. 62073159 and Grant No. 62003155, in part by  GHP/126/21GD from Innovation and Techology Commission, Hong Kong, in part by the Shenzhen Science and Technology Program under Grant No. JCYJ20200109141601708, and in part by the Science, Technology and Innovation Commission of Shenzhen Municipality under Grant No. ZDSYS20200811143601004.}
}

\begin{document}
\maketitle

\begin{abstract}
In robotic insertion tasks where the uncertainty exceeds the allowable tolerance, a good search strategy is essential for successful insertion and significantly influences efficiency. The commonly used blind search method is time-consuming and does not exploit the rich contact information. In this paper, we propose a novel search strategy that actively utilizes the information contained in the contact configuration and shows high efficiency. In particular, we formulate this problem as a Partially Observable Markov Decision Process (POMDP) with carefully designed primitives based on an in-depth analysis of the contact configuration's static stability. From the formulated POMDP, we can derive a novel search strategy. Thanks to its simplicity, this search strategy can be incorporated into a Finite-State-Machine (FSM) controller. The behaviors of the FSM controller are realized through a low-level Cartesian Impedance Controller. Our method is based purely on the robot's proprioceptive sensing and does not need visual or tactile sensors. To evaluate the effectiveness of our proposed strategy and control framework, we conduct extensive comparison experiments in simulation, where we compare our method with the baseline approach. The results demonstrate that our proposed method achieves a higher success rate with a shorter search time and search trajectory length compared to the baseline method. Additionally, we show that our method is robust to various initial displacement errors.

\end{abstract}

\section{Introduction}\label{sec:intro}
The insertion task, which involves mating different components of a product, is one of the most common tasks for industrial robots in the manufacturing industry. Despite being basic, its performance significantly influences product quality and production efficiency. However, the insertion task is challenging because of its contact-rich nature and high precision requirement. Due to these challenges, the robotic insertion problem has attracted considerable research attention in recent decades~\cite{Whitney1982Quasi, simunovia1979information, drake1978using, Lozano-Perez1984IJRRAutomatic, Chhatpar2001IROS, Triyonoputro2019, Song2014IROS, Lian2021IROSBenchmarking, LuoRSS21Robust}. Generally speaking, the insertion task involves four phases: grasp, approach, search, and insert. During an insertion task, the robot first \textit{grasps} the peg and \textit{approaches} the corresponding hole; then, it lines up the peg with the hole using a \textit{search} strategy and \textit{inserts} the peg. Among these phases, the search phase is usually the most time-consuming one~\cite{Triyonoputro2019, Lian2021IROSBenchmarking}. Even though the search phase can be omitted with compliance for simple scenarios in which the error tolerance is higher than the uncertainty \cite{Whitney1982Quasi}, a good search strategy is vital for tasks requiring high precision (\eg, chamferless holes) and tasks involving high uncertainty (\eg, household robots)~\cite{Chhatpar2001IROS, Triyonoputro2019, Lian2021IROSBenchmarking}.

A commonly used search strategy is blind search~\cite{Chhatpar2001IROS, Lian2021IROSBenchmarking, Haugaard2021CORL}, which involves moving the peg along a predefined path (\eg, spiral or Lissajous curve) while pushing it against the hole's surface. The search continues until the peg is aligned with the hole. This method guarantees locating the hole if the search path is dense enough. However, it is inefficient because the robot might need to move the peg for a long distance, depending on the initial positional error. 

\begin{figure}[t]
    \centering
     \includegraphics[width=\linewidth]{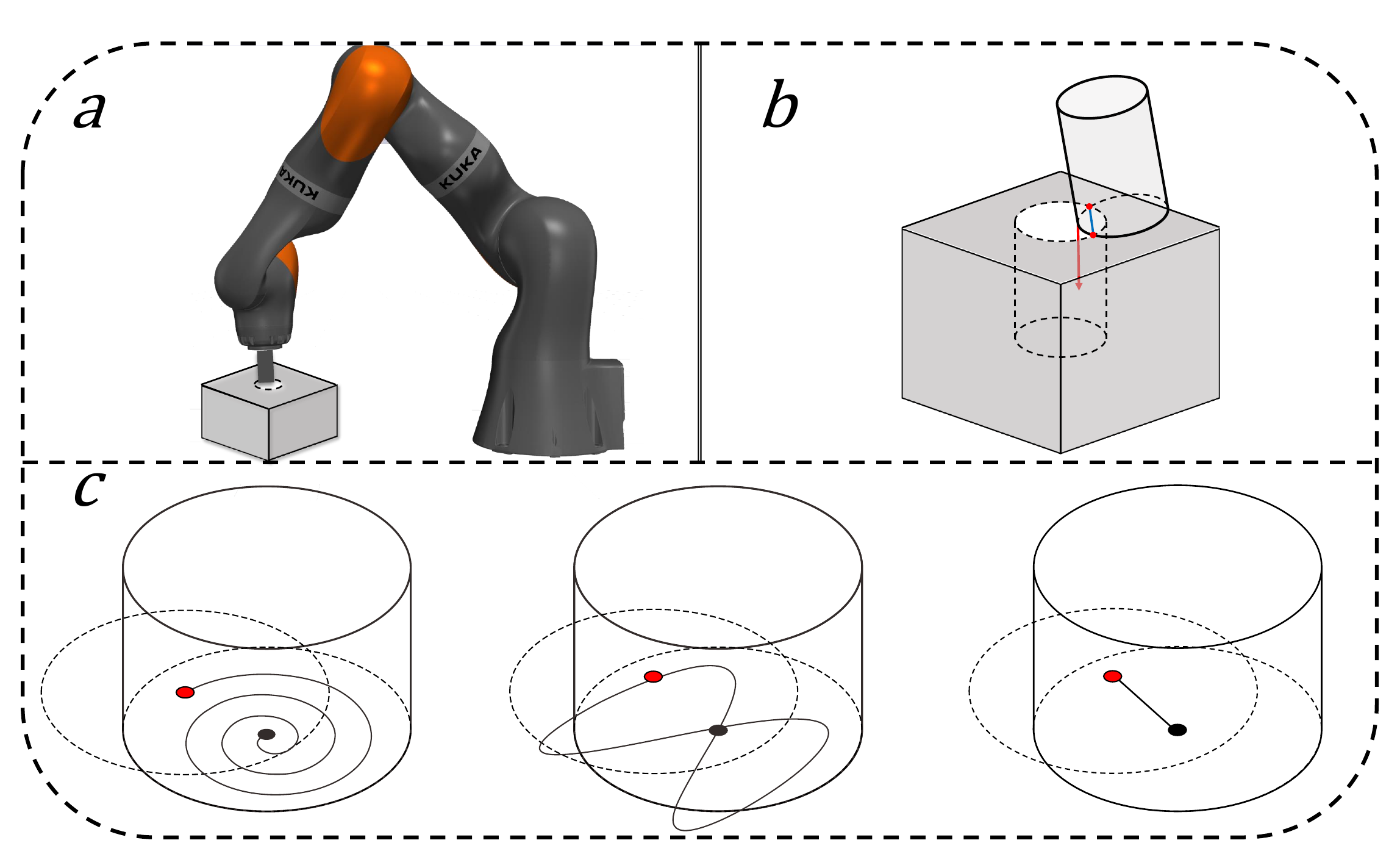}
    \caption{\textbf{An overview of the proposed method}.
(a) Illustration of the search problem. The robot needs to align the peg with the hole. (b) Illustration of the proposed search strategy. Our search strategy actively applies force to exploit the information contained in the contact configuration and locate the hole's position. (c) Search paths of different search methods. From left to right: blind search with Spiral curve, blind search with Lissajous curve, and our method. The hole and peg centers are represented as black and red circles, respectively. Our search method can directly align the peg with the hole along a straight line.}
% During the search phase, the robot needs to align the peg with the hole. The red circle represents the hole's center, and the black circle represents the peg's end's center. The black curve represents the path the peg follows during the search phase. From left to right: blind search with Sprial curve, blind search with Lissajous curve, and our proposed search strategy. Our search strategy can directly align the peg with the hole along a straight line.}
    \label{fig:cover_picture}
    \vspace{-5mm}
\end{figure}

In contrast, when humans unlock a door with a key in the dark, they are unlikely to follow a predefined path. Instead, they may use a more active strategy based on haptic sensing. Before actually aligning the key with the keyhole, they have an estimation of the hole's position. This example highlights one missing element in the blind search: the contact configuration between the peg and the hole, which contains rich information about relative displacement. How to make the search process more intelligent has attracted a lot of research attention~\cite{Newman2001ICRA, Kim2012CASE, Kim2022ICRAActive, Haugaard2021CORL, Song2014IROS}.

This paper focuses on the search problem and proposes an efficient search strategy that makes use of contact information between the peg and the hole, leveraging only the robot's proprioceptive sensory information. Specifically, we first formulate the search problem as \textit{Partially Observable Markov Decision Process} (POMDP) and design its primitives by analyzing the static stability of the contact configuration. Then, we derive a search strategy from the formulated POMDP and incorporate the strategy to construct a Finite-State-Machine (FSM) controller for the insertion task. In particular, to achieve the FSM controller's behaviors, we use a low-level Cartesian Impedance Controller. The effectiveness of our method is demonstrated through comprehensive comparison experiments with the baseline method.

\subsection{Related Works}

\textit{\textbf{Insertion Phase.}}
Robotic insertion has been studied for several decades due to its great importance. Early works \cite{Whitney1982Quasi, simunovia1979information, Caine1989ICRA}
analyze the contact wrenches acting on the peg when the peg has been partially inserted into the hole, and identify two important failure modes, jamming and wedging. These works provide guidance in the design of the robotic system's compliance for insertion task, which leads to the hardware remote-center-compliance (RCC) \cite{drake1978using}. RCC responds quickly to external forces but lacks flexibility for different applications \cite{villani2016force}. To deal with this problem, Cartesian Impedance Control~\cite{Hogan1985Impedance} is proposed to achieve compliance through software control and has been widely used in the robotic insertion task~\cite{Caccavale1998Control, Song2014IROS, Johannsmeier2019ICRA, Nigro2020IROS, Korbinian2020ICRARobust}. It can compensate for the positional error and help to avoid jamming and wedging modes.

\textit{\textbf{Search Phase.}} Another concentration on the robotic insertion community is in the search phase. That is, given a rough initial estimation of the hole's pose, how to align the peg with the hole before the insertion phase. Search with a predefined path is a widely adopted method \cite{Chhatpar2001IROS} and has shown effectiveness in reducing uncertainty. To make the search process more intelligent, strategies based on force \cite{Newman2001ICRA, Kim2012CASE, Korbinian2020ICRARobust, Kim2022ICRAActive} or vision sensors \cite{Song2014IROS, Nigro2020IROS, Haugaard2021CORL, Gao2021RAL, Morgan2021RSSVisionDriven} are proposed. As for the force sensors, in \cite{Newman2001ICRA}, the reaction moment peaks during two scan motions are used to interpret the hole's coordinates. Later in \cite{Kim2012CASE}, the authors propose a hole detection algorithm based on the measured external wrench, which requires an F-T sensor with high precision installed on the end-effector. In \cite{Korbinian2020ICRARobust}, the authors propose a particle filter-based algorithm that updates the belief of the hole's position based on the robot's proprioceptive sensing. Nevertheless, the search phase in this work still depends on the Lissajous search. In \cite{Kim2022ICRAActive}, the authors also tried to estimate the contact configuration between the peg and the hole, but their method requires a tactile sensor that provides more information. Compared with the existing methods, our method does not require tactile information and is based purely on the robot's proprioceptive sensing ability. The length of the search trajectory in our method is also minimized. There are also works that try to estimate the hole's pose with vision directly. In \cite{Song2014IROS}, the template-based matching method is used to estimate the hole's position. The more recent works \cite{Haugaard2021CORL, Gao2021RAL, Morgan2021RSSVisionDriven} use a learned neural network to predict the hole's relative pose. However, the vision method is subject to occlusion problems, and the prediction accuracy is not enough for scenarios demanding high precision in general. 

\textit{\textbf{POMDP and Reinforcement Learning.}} Some recent works formulate the whole robotic insertion task as POMDP problems. Nonetheless, the formulated POMDP problems are generally intractable to solve directly, so reinforcement learning is adopted \cite{Inoue2017Iros, Schoettler2020IROS, Lee2020TROMaking, WirnshoferRSS20Controlling, LuoRSS21Robust}. Reinforcement learning learns a policy by interacting with the environment in a trial-and-error manner.  The learned policy has the advantage of directly mapping multi-modal sensing data into action~\cite{Lee2020TROMaking, LuoRSS21Robust}. However, reinforcement learning is low in data efficiency, and a large amount of data is required to train the agent. Moreover, the agent trained in the simulation environment also suffers from the sim-to-real gap problem. 

\subsection{Contribution}
The contributions of this paper are as follows. First, we offer a novel and systematic perspective on the search problem through a POMDP formulation. 
By leveraging static stability analysis to design the primitives, we achieve a significant advantage in consistently and accurately identifying the hole's location.
% First, we provide a novel and systematic formulation of the search problem as a POMDP. We leverage static stability analysis to create primitives for the POMDP, which allows us to consistently and accurately pinpoint the hole's location. 
% Second, building upon the POMDP, we develop a comprehensive and effective control framework, which efficiently solves the robotic insertion task.
% Second, building upon the POMDP, we efficiently solve the robotic insertion task by developing a comprehensive and effective control framework.
Second, built upon the formulated POMDP, we propose a novel search strategy and incorporate it into a comprehensive control framework, which efficiently solves the robotic insertion task. 
%, which efficiently solves the search problem. 
Third, our extensive experiments show that the proposed method outperforms the baseline approach in terms of success rate, search times, and trajectory lengths. Furthermore, our approach is notably robust to various initial displacement errors.
% Second, built upon the formulated POMDP, we propose a novel search strategy. With this search strategy, we develop a comprehensive and effective control framework, which solves the robotic insertion task efficiently.  

% propose a novel search strategy that can efficiently align the peg with the hole using proprioceptive sensing only. Second, our approach provides a systematic formulation of the search problem using POMDP. We leverage static stability analysis to create primitives for the POMDP, which allows us to consistently and accurately pinpoint the location of the hole. Third, our extensive experiments show that the proposed method outperforms the baseline approach in terms of success rate, search times, and trajectory lengths. Furthermore, our approach is notably robust to various initial displacement errors.

\section{Problem Description}\label{sec:prob_des}
\begin{figure}[b]
    \begin{subfigure}[b]{.15\textwidth}
    \centering
    \includegraphics[width=\textwidth]{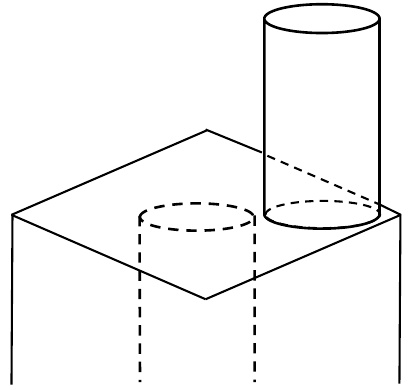}
    \caption{}
    \end{subfigure}
    \hfill
    \begin{subfigure}[b]{.15\textwidth}
    \centering
    \includegraphics[width=\textwidth]{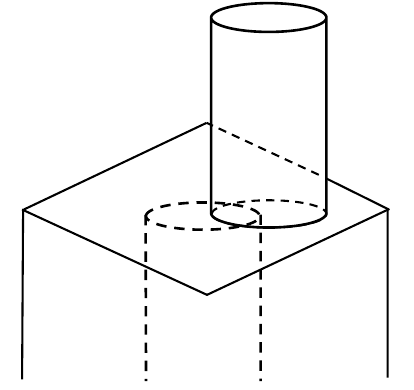}
    \caption{}
    \label{fig:partially_overlap}
    \end{subfigure}
    \hfill
    \begin{subfigure}[b]{0.15\textwidth}
    \centering
    \includegraphics[width=\textwidth]{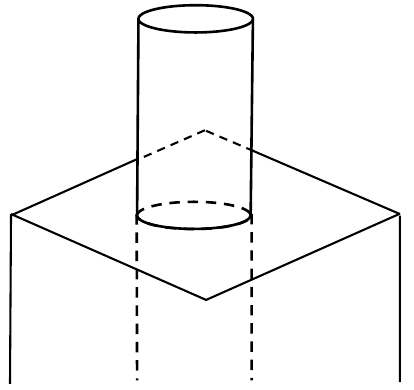}
    \caption{}
    \end{subfigure}
    \caption{\textbf{Three types of contact configuration between peg and hole.} (a) No overlapping (b) Partially overlapping (c) Perfect alignment}
    \label{fig:3TypesContactScenarios}
    % \vspace{-15px}
\end{figure}

In this work, we focus on the search phase of the robotic insertion task. During the search phase, the robot needs to reduce the uncertainty presented in the initial estimated hole's position. In particular, we consider a torque-controlled robotic arm with a round peg rigidly attached to its end, and the peg's relative pose with the robot's end-effector is known. The robot can be controlled by commanding torque $\commandTorque$, and the dynamics equation that governs the arm's motion is: 
\begin{equation}
   \mass(\q)\ddq + \coriolis(\q, \dq)\dq + \gravityTorque(\q) = \commandTorque + \sum_{i} \jacobian^T_i(\q) \wrench_{i}^\text{ext}
   \label{eq:manipulatorEquation}
\end{equation}
where $\q \in \R^n$ is the joint position, $\dq \in \R^n$ is the joint velocity, $\mass(\q) \in \R^{n\times n}$ is the mass matrix, $\coriolis(\q, \dq)$ summarizes the centripetal and Coriolis effects, $\gravityTorque$ is the torque on the robot caused by gravity and $\jacobian^T_i(q)\wrench_{i}^\text{ext}$ is the torque due to the $i$-th external contact. The total external contact torque is $\externalTau = \sum_{i} \jacobian^T_i(\q) \wrench_{i}^\text{ext}$, and $\jacobian_i(\q)$ is the $i$-th contact jacobian. The robot can sense its joint position $\q$, joint velocity $\dq$, and the external torque $\tau_{ext}$ acting on it. We define the plan of the robot as a feedback control law $\commandTorque(\q, \dq, \tau_\text{ext}, t)$ that will generate a sequence of motions of the robot. The frames that we are interested in are the world frame $\worldFrame$, the peg frame $\pegFrame$, and the hole frame $\holeFrame$, as shown in Fig.\ref{fig:simulation_setup}. The world frame $\worldFrame$ locates at the robot's base, the peg frame $\pegFrame$ locates at the center of the peg's end, and the hole frame $\holeFrame$ locates at the center of the hole's top. The pose of the peg in world frame $^W T^P$ can be obtained with the robot's forward kinematics:
\begin{equation}
    ^W T ^P = \phi(\q)
    \label{eq:forward_kinematics}
\end{equation}

Initially, a rough estimation of the hole's position $^W \hat{p}^H = (\hat{x}, \hat{y}, \hat{z})$ is given to the robot.  At the beginning of the search phase, there are three types of contact configuration between the peg and the hole, as shown in Fig.~\ref{fig:3TypesContactScenarios}: 
\begin{itemize}
    \item \textit{No overlapping}: the peg does not overlap with the hole.
    \item \textit{Partially overlapping}: the peg partially overlaps the hole.
    \item \textit{Perfect Alignment} :The peg is aligned with the hole.
\end{itemize}
In this paper, we consider that at the beginning of the search phase, the peg partially overlaps the hole (see Fig.~\ref{fig:partially_overlap}). 
\begin{remark}
For the \textit{no overlapping} case, the contact configuration contains no information about the hole's relative position. For the \textit{perfect alignment} case, the peg is already aligned with the hole, and the search phase is unnecessary. It should be noted that in reality, the estimated hole's position $^W \hat{p} ^H$ is neither accurate enough to directly align the peg with the hole nor too noisy that the hole is completely missed. Therefore, we focus on the \textit{Partially overlapping} case.
\end{remark}

Because of the uncertainty in $^W \hat{p} ^H$, there would be a displacement error between the peg and the hole $^P p ^H$, which can be further reduced to a 2D vector $\displacement = (\delta x, \delta y)$ in $x$- and $y$- direction. During the search phase, the robot needs to move the peg and reduce the displacement error $\displacement$ to a goal set $\goalSet$. The goal set $\goalSet$ is determined by the hole's clearance and consists of displacements $\displacement$ with which the peg is aligned. Therefore, the search problem during the search phase can be formulated as:
\begin{problem}[Search Problem]
\label{def:search_problem}
Given an initial displacement $\displacement(0) = (\delta x_0, \delta y_0)$ between the peg and the hole, find a plan $\commandTorque(\q, \dq, \tau_\text{ext}, t)$ of the robot, such that there exists a time $T < \infty$, $\displacement(t) \in \goalSet$ for $t > T$. 
\end{problem}

\section{POMDP formulation}
As introduced before, the blind search method neglects the information contained in the contact configuration of the peg and the hole. In order to utilize the information contained in the contact configuration, the robot needs to actively interact with the environment to collect information, estimate the displacement and move the peg accordingly. The POMDP models an agent interacting synchronously with a world and trying to optimize a reward. The agent does not have direct access to the world's states and can only observe some intermediate observations. It needs to maintain a belief about the current state and update the belief based on the observation. In order to optimize the reward, the agent chooses actions to collect information and change the world state. In our search problem~\ref{def:search_problem}, the actual location of the hole's position is unknown and active interaction with the environment is necessary. Therefore, we model Problem.~\ref{def:search_problem} as a POMDP. In particular, we tackle Problem.~\ref{def:search_problem} in a hierarchical manner. In the formulated POMDP problem, we omit details of the robot's dynamics~\eqref{eq:manipulatorEquation}. The solution to the formulated POMDP problem is a sequence of action primitives, which is then converted to the robot's plan $\commandTorque(\q, \dq, \tau_\text{ext}, t)$ by a low-level controller.

The formulated POMDP problem $\POMDP$ can be represented with a tuple $(\stateSpace, \actionSpace, \observationSpace, \transitionModel, \observationModel, \rewardFunction)$, where $\stateSpace$ is the state space, $\actionSpace$ is the action space, $\observationSpace$ is the observation space, $\transitionModel$ is the transition model, $\observationModel$ is the observation model, and $\rewardFunction$ is the reward function. The relevant state in our problem is $s = (\displacement)$, which is the displacement $\displacement$. The action space $\actionSpace = \{a_i\}$ is a set of action primitives that are used to collect information or move the peg. The observation space $\observationSpace = \{o_i\}$ is a set of observation primitives that summarize high-level information. The transition model $\transitionModel(s, a, s') = p(s' | s, a)$ and the observation model $\Omega(o, s, a) = p (o | s, a)$ are based on the primitives and the contact model. In particular, we model $\transitionModel$ and $\observationModel$ as deterministic models, and the only uncertainty of the formulated POMDP problem $\POMDP$ is in the displacement $\displacement$, which is not directly observed.  The robot keeps an internal belief of $v$, $\belief(v) \in [0, 1]$, with $\int \belief(\displacement) d\displacement = 1$. 
The initial belief $\belief_0(\displacement)$ is a uniform distribution on $\displacementSet$:
\begin{equation}
    \belief_0(\displacement) = 
    \begin{cases}
        1 / \int_\displacementSet 1 d \displacement,  &\displacement \in \displacementSet \\
        0, &\displacement \notin \displacementSet
    \end{cases}
\end{equation}
where $\displacementSet$ is the set of $\displacement$ that the peg partially overlaps the hole. To describe the goal of aligning the peg with the hole, the reward function is defined as:
\begin{equation}
    \rewardFunction(s, a) = 
    \begin{cases}
        1, & \displacement \in \goalSet  \\
        -1, & \text{otherwise}
    \end{cases}
\end{equation}

In the formulated POMDP $\POMDP$, we use action primitives $a_i$ to narrow the action space. The action primitives $a_i$ are high-level motion plans of the robots. The usage of action primitives naturally decomposes the original problem of finding robot plan $\commandTorque(\q, \dq, \tau_\text{ext}, t)$ in a hierarchical manner: 1) finding a sequence of action primitives that achieve the goal; 2) realization of each action primitive with the commanded torque $\commandTorque$ by a low-level controller. We also design observation primitives $o_i$ that summarize the high-level information of possible outcomes. The usage of these primitives simplifies the POMDP $\POMDP$ and makes the planning feasible. However, careful design of these primitives is necessary, especially for collecting and interpreting meaningful information to update the belief $\belief$. Our design of these primitives is based on the observation that the static stability of the peg depends on both the applied force on the peg and the contact configuration. The reaction of the peg with respect to the applied force can be used to collect information about $\displacement$. What's more, the outcome of our designed action primitives is reliable. This enables us to have deterministic transition and observation models, $\transitionModel(s, a, s')$ and $\observationModel(o, s, a)$. The belief can thus be updated determinately, and the solution to the POMDP $\POMDP$ can be found directly.

\section{Action and Observation Primitives Design}
To illustrate the idea of our designed primitives, we will start with the simplified planar search problem for which a graphical analysis of static stability is possible. The designed action and observation primitives in the planar case can be generalized to the 3D case.

\subsection{The simplified planar problem}

\begin{figure}[htbp]
    \centering
    \includegraphics[width=\linewidth]{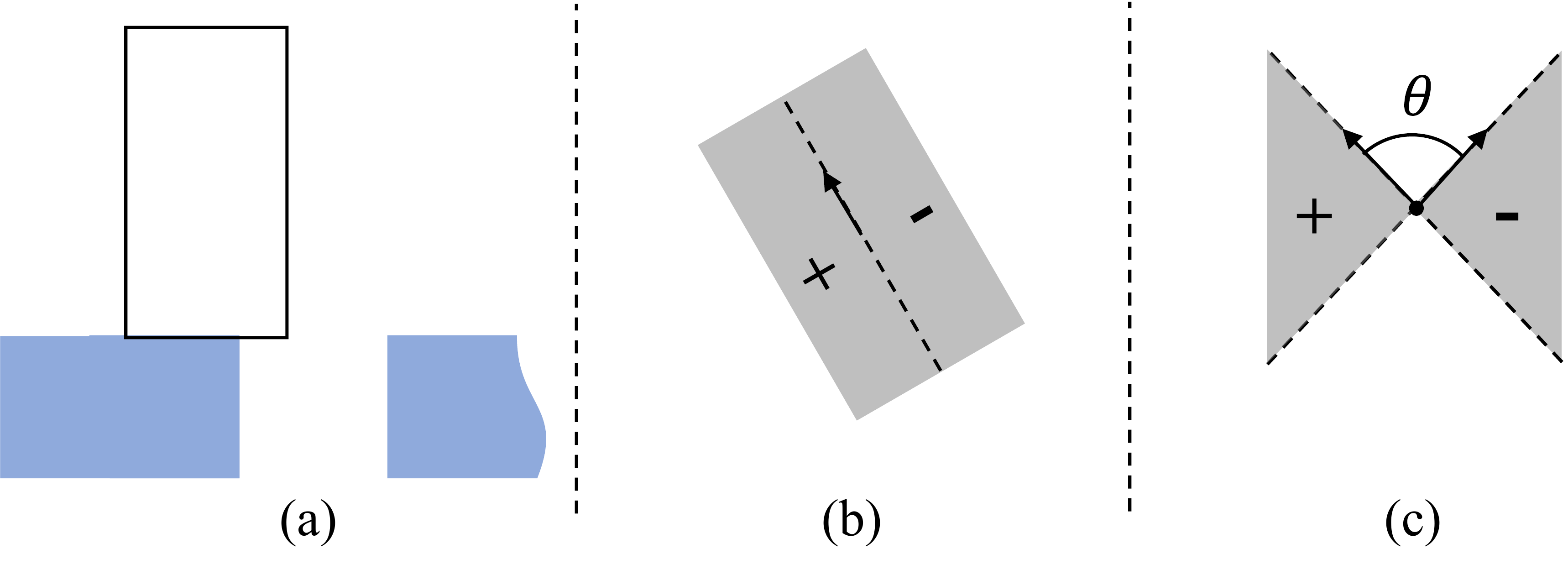}
    \caption{\textbf{The planar problem and moment labeling.} (a) The simplified planar problem. (b) Moment labeling of a contact wrench. (c) Moment labeling of a friction contact, where $\theta = 2\arctan(\mu)$ and $\mu$ is the friction coefficient.
    The ``+'' region consists of points where all contact wrenches only apply positive moment, the ``-'' region consists of points where all contact wrenches only apply negative moment, and the white region consists of points where the contact wrenches can apply both positive and negative moment.}
    \label{fig:MomentLabel}
    \vspace{-6px}
\end{figure}

The simplified planar problem is shown in Fig.~\ref{fig:MomentLabel}a. For the planar problem, we can use moment labeling \cite{Mason1991Two, lynch2017modern} to analyze the stability of the peg with respect to different applied forces, which facilitates our design of primitives. The moment labeling is a graphical way to represent the planar contact wrenches. Illustrations of moment labeling are shown in Fig.~\ref{fig:MomentLabel}b and Fig.~\ref{fig:MomentLabel}c. 

\begin{figure}[htbp]
    \centering
    \includegraphics[width=\linewidth]{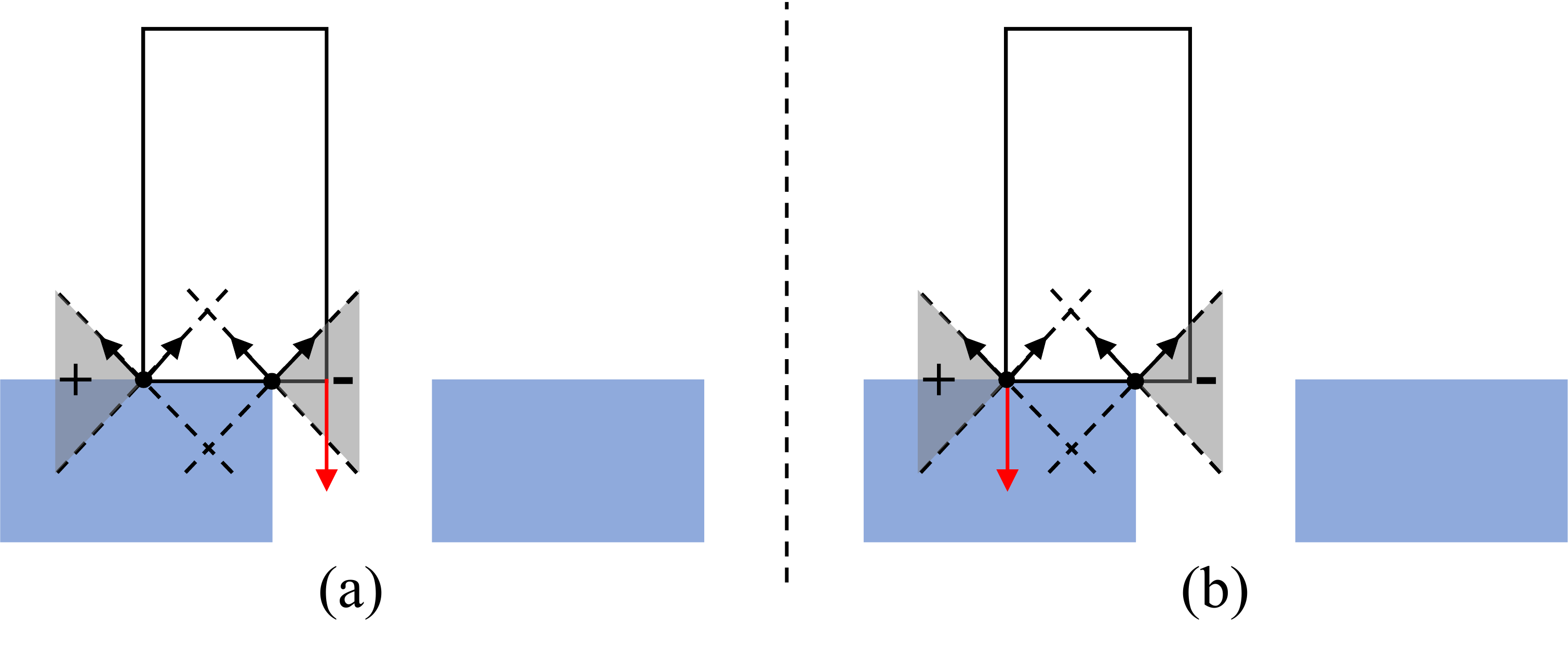}
    \caption{\textbf{Moment labeling analysis of the planar problem.} The red arrow represents the applied force on the peg. (a) Apply a downward force at the right end of the peg. The applied force cannot be compensated by the contact wrenches. The peg will tilt rightward. (b) Apply a downward force at the left end of the peg. The applied force can be compensated by the contact wrenches. The peg will keep static.}
    \label{fig:PlanarProblem}
    \vspace{-15px}
\end{figure}

The moment labeling of the planar peg in contact with the hole is shown in Fig.~\ref{fig:PlanarProblem}. Using moment labeling, we can analyze the static stability of the peg with respect to different applied forces. As can be seen from Fig.~\ref{fig:PlanarProblem}a, if there is a downward applied force at the right end of the peg, the peg will tilt rightward because the contact wrenches cannot compensate for the applied force. On the contrary, if the applied downward force is at the left end of the peg, the applied force can be compensated by the contact wrenches, and the peg will stay static, as shown in Fig.~\ref{fig:PlanarProblem}b. Similar logic can be applied when the peg is on the right side of the hole. By exploiting this static stability analysis, we can design two action primitives for collecting information:
\begin{itemize}
    \item $a_{fl}$: apply a downward force at the left end of the peg.
    \item $a_{fr}$: apply a downward force at the right end of the peg.
\end{itemize}

Together with the movement action primitives, the designed action primitives are $a\in\{a_{fl}, a_{fr}, a_{ml}, a_{mr}, a_{s}\}$, where $a_{ml}$ is moving the peg leftward for time $T_a$, $a_{mr}$ is moving the peg rightward for time $T_a$ and $a_{s}$ is stay in current position. The corresponding observation primitives are $o \in \{o_s, o_l, o_r, o_i\}$, where $o_s$ is static, $o_l$ is tilting leftward, $o_r$ is tilting rightward, and $o_i$ is peg being partly inserted into the hole. The belief $\belief$ can also be simplified to be on 3 symbolic states:$\phi \in \{h_l, h_r, h_a\}$, where $h_l$, $h_r$ and $h_a$ means the hole is on the left side, on the right side and aligned, respectively. The initial belief on these symbolic states is $\{0.5, 0.5, 0\}$. With the designed action and observation primitives, we can model the transition model $\transitionModel$ and the observation model $\observationModel$ as deterministic. Therefore, the belief $\belief$ can be updated deterministically with observed $o$. For example, if the observation $o_l$ is observed, the belief can be updated to $\{1.0, 0, 0\}$.

Because of the simplicity of the formulated POMDP problem $\POMDP$, we can give a search strategy for the planar problem: the robot firstly applies $a_{fl}$ or $a_{fr}$; based on the received observation $o$, the belief $b$ is then updated; then, the robot continually applies $a_{ml}$ or $a_{mr}$ according to $\belief$  until $o_{i}$ is observed.

\subsection{The 3D problem}
\begin{figure}[htbp]
    \vspace{-4mm}
    \begin{subfigure}[b]{.24\textwidth}
    \centering
    \includegraphics[width=\textwidth]{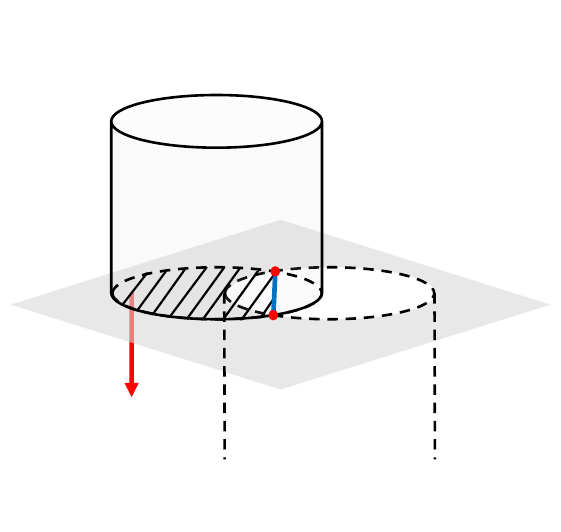}
    \vspace{-10mm}
    \caption{}
    \end{subfigure}
    \hfill
    \begin{subfigure}[b]{.24\textwidth}
    \centering
    \includegraphics[width=\textwidth]{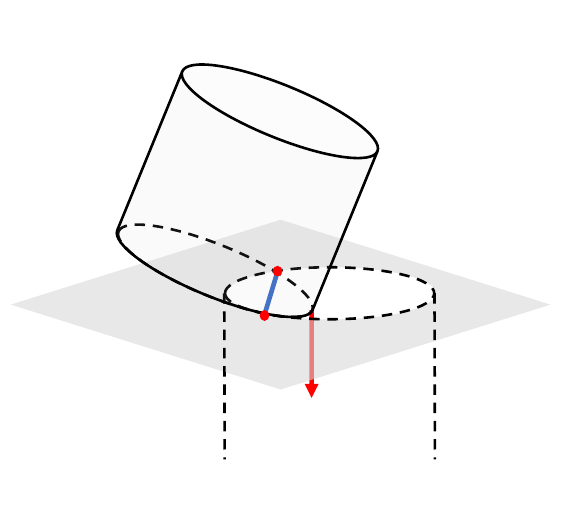}
    \vspace{-10mm}
    \caption{}
    \end{subfigure}
    \caption{\textbf{Static stability analysis of the 3D contact configuration between the peg and the hole.} The support polygon is the shadowed region. The intersection line is the blue line with two red-point ends. (a) If the projection of the downward applied force is inside the support polygon, the peg will keep static. (b) If the projection of the downward applied force is outside the support polygon, the peg will tilt around the intersection line.}
    \label{fig:3DPeg}
    \vspace{-15px}
\end{figure}

It turns out that the above idea of designing primitives based on the static stability of the peg with the applied force can be generalized to the 3D case. To simplify the static stability analysis, we use the idea of support polygon \cite{McGhee1968OnTheStability, Bretl2008TestingStatic} from the legged robot community. The support polygon is the convex hull of the legged robot's all contact points with the flat ground. It is used to characterize the static stability of the legged robot: if the projection of the legged robot's center of mass (COM) to the ground is inside the support polygon, the robot is stable; otherwise, if the projection is outside the support polygon, the robot will tilt. The static stability analysis of the 3D peg can be done in a similar way. For our 3D partially aligned peg, the support polygon is the convex hull of all the contact points between the peg and the hole, as shown in Fig.~\ref{fig:3DPeg}a. If the projection of the applied downward force to the horizontal plane is inside the support polygon, as shown in Fig.~\ref{fig:3DPeg}a, then the peg will keep static; otherwise, the peg will tilt, as shown in Fig.~\ref{fig:3DPeg}b. 

Similar to the planar case, if we choose the downward applied force inside the peg's end, we can be sure that if the peg tilts, it tilts towards the hole. Therefore, we can design the action primitives for collecting information as $\{a_f^1, a_f^2, \dots, a_f^K\}$, where $a_f^i$ applies downward force at different locations inside the peg's end. It should be noted that, unlike the planar case, in 3D, two action primitives are not enough for collecting information, as they might not lie outside the support polygon to tilt the peg. The downward applied forces should be uniformly distributed around the perimeter of the peg end, and $K$ should be chosen large enough to cover a sufficient portion of the perimeter.
The movement action primitives are designed as continuous $\{a_m = (\hat{n}, T_a)\}$, which is moving the peg in a 2D direction $\hat{n}$ for time $T_a$. The total action primitives are $\{a_f^1, a_f^2, \dots, a_f^K\} \cup \{a_m, a_s\}$. 

In the planar case, the observation primitives contain $o_s$ and $o_i$. However, the two observation primitives $o_l$ and $o_r$ are not enough to characterize the tilting motions in 3D. By assuming that the friction is large enough such that sliding does not happen when the peg tilts with action $a_f^i$, the tilting motion is a rotation around the intersection line between the peg and the hole. What's more, this intersection line can be calculated based on the trajectory of the peg frame $\pegFrame$'s origin $^W p ^P(t)$, which can be obtained based on the forward kinematics~\eqref{eq:forward_kinematics}. Therefore, we design the observation primitives characterizing the tilting motion as $\{o_t = (\hat{k}, p)\}$, where $\hat{k}\in \R^2$ is the direction of the intersection line and $p\in \R^2$ is a point that this intersection line passes. The total observation primitives are $\{o_s, o_i\} \cup \{o_t\}$, which summarizes the possible outcomes of the designed action primitives.

The observation primitive $o_t$ contains information on the contact configuration. Similar to the planar case, we model $\transitionModel$ and $\observationModel$ as deterministic. The belief $\belief(\displacement)$ 
can be updated with $o_t$ determinately based on geometric analysis. Suppose that $(\hat{k}, p)$ is observed for a round peg with radius $r$,
based on symmetry, it immediately follows that $\displacement$ is perpendicular to $\hat{k}$ and the direction of $\displacement$ is $\rotationMatrix(-\frac{\pi}{2})\hat{k}$, where 
\begin{equation}
    \rotationMatrix(\theta) = \begin{bmatrix}
    \cos(\theta) & -\sin(\theta) \\
    \sin(\theta) & \cos(\theta) \\
    \end{bmatrix}
\end{equation}
What's more, with $p$ known, the length $l$ of the intersection line can be obtained. The belief $\belief(\displacement)$ can then be updated into a Dirac delta function 
\begin{equation}
    \belief(\displacement) = \begin{cases}
    \infty, &\displacement = 2\sqrt{r^2 - l^2 / 4}\rotationMatrix(-\pi/2)\hat{k} \\
    0, & \text{otherwise}\\
    \end{cases}
\end{equation}
with $\int_\displacement \belief(\displacement) d\displacement = 1$.

Like the planar problem, because of the simplicity  brought by the designed primitives, we can give an efficient search strategy to the formulated POMDP problem $\POMDP$. The robot firstly use actions $a_f^1$ to $a_f^K$ until $o_t = (\hat{k}, p)$ is observed. With $o_t$, the robot updates the belief $\belief$ accordingly. Then based on the updated belief, the robot continually moves the peg towards the hole using $a_m = (\rotationMatrix(-\frac{\pi}{2})\hat{k}, T_a)$ until $o_i$ is observed.

\section{Control Framework}
\label{sec:fsm}
\begin{figure}[htbp]
    \centering
    \includegraphics[width=0.92\linewidth]{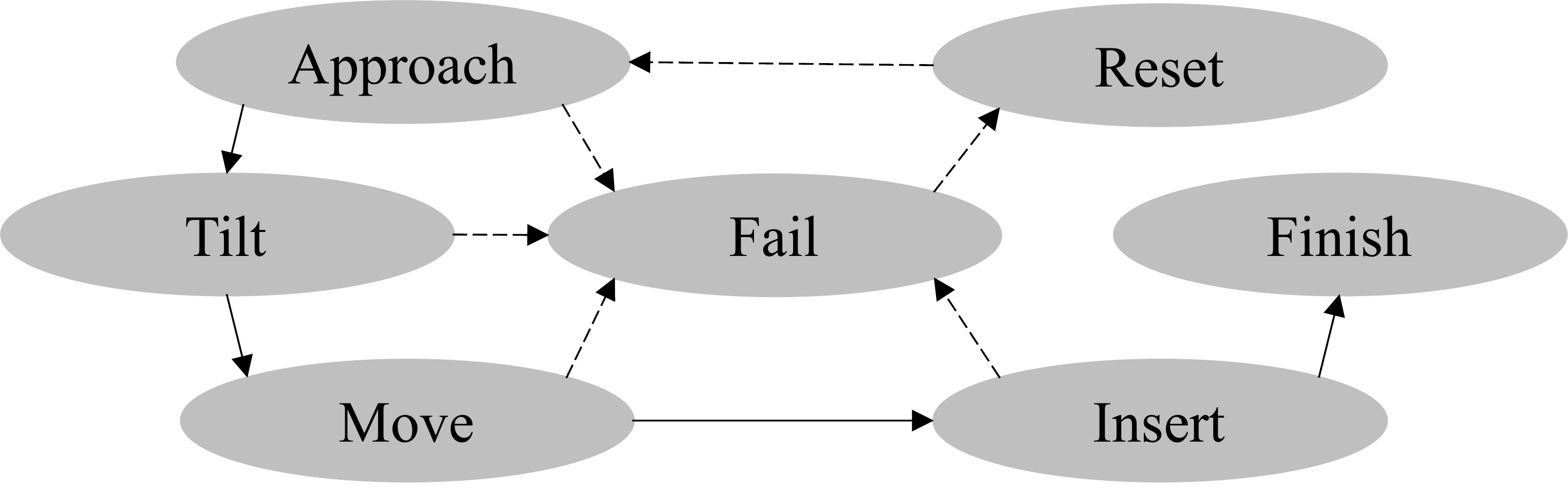}
    \caption{\textbf{State transitions of the proposed FSM controller.} The transitions in a normal insertion process are shown as solid arrows. A reset process is triggered if the robot fails to complete the insertion successfully. The transitions in the reset process are shown as dashed arrows.}
    \label{fig:fsm}
\end{figure}

The solution to the formulated POMDP problem is an efficient search strategy consisting of a sequence of action primitives conditioned on the observation primitives. Because of the simplicity of our proposed search strategy, we can incorporate it into a Finite-State-Machine (FSM) controller for completing the robotic insertion task. The FSM controller needs to complete the peg-in-hole task from the \textit{approach} phase to the \textit{insert} phase. The states and the associated robot behaviors of the proposed FSM controller are:
\begin{enumerate}
    \item \textit{Approach}: The robot moves the peg down to approach the hole.
    \item \textit{Tilt}: The robot applies action primitives $a_f^1$ to $a_f^K$ in sequence.
    \item \textit{Move}: The robot applies action primitive $a_m$ based on the belief $\belief(\displacement)$.
    \item \textit{Insert}: The robot inserts the peg into the hole.
    \item \textit{Finish}: The robot has successfully inserted the peg.
    \item \textit{Fail}: The robot fails in previous states.
    \item \textit{Reset}: The robot moves to the start position.
\end{enumerate}
The search strategy is implemented in states \textit{Tilt} and \textit{Move}. The FSM controller is illustrated in Fig.~\ref{fig:fsm}. The transition conditions between different states are manually designed and are based on $\q, \dq$, and $\externalTau$.

In order to convert the behaviors in the FSM controller to the robot's plan $\commandTorque(\q, \dq, \tau_\text{ext}, t)$, a low-level controller is needed. During the peg-in-hole task, the robot will inevitably make contact with the environment. To make the robot's behaviors in the presence of contact robust, we use the Cartesian Impedance Control as the low-level controller. The pose of the peg $^WT^P$ can be expressed as a 6-dimension vector $x_p = (^Wp^P, \varphi_p)$ where $^Wp^P$ is the peg frame's position and $\varphi_p$ is the peg frame's orientation expressed in Euler angle. Denote $x_d$ as the desired pose of the peg frame and
\begin{equation}
    A(\varphi_p) =
    \begin{bmatrix}
        I & 0 \\
        0 & T(\varphi_p)
    \end{bmatrix}
\end{equation}
where $T(\varphi_p)$ serves to map the derivative of the Euler angle to the angular velocity. Then, the commanded torque $\commandTorque$ in~\eqref{eq:manipulatorEquation} to achieve the Cartesian Impedance Control is:
\begin{multline}
   \commandTorque = \gravityTorque(q) + \coriolis(\q, \dq)\dq + \feedforwardTau  \\ 
    + \jacobian_p^T(\q)[A^{-T}(\varphi)K_p(x_d - x_p) - K_d(v_d - v_p)] \\ 
    + (I - \jacobian_p^{\dag}J_p)[K_p^{null}(\q_0 - \q) - K_d^{null}(\dq)]
\end{multline}
where $\jacobian_p(q)$ is the robot's jacobian at the peg's frame and $\jacobian_p^{\dag}$ is the jacobian's pseudo-inverse; $K_p, K_d \in \R^{6\times6}$ is the gain matrix in the Cartesian space; $v_p$ and $v_d$ is the real and desired twist of the peg's frame, respectively; $K_p^{null}, K_d^{null} \in \R^{n\times n}$ is the gain matrix to stabilize the controller in the null space; $\q_0$ is the robot's normal position; $\feedforwardTau$ is the additional feedforward torque. For more details about the Cartesian Impedance Control, the readers are recommended to refer to~\cite{villani2016force}. It should be noted that in $\commandTorque$, we have an additional feed-forward term $\feedforwardTau$ to apply the required wrench, as used in action primitive $a_f^i$. Suppose that we would like to apply wrench $F$ in frame $\{a\}$, then  
\begin{equation}
\feedforwardTau = \jacobian^T_a(\q) F 
\end{equation}
where $\jacobian_a$ is the robot's jacobian at frame $\{a\}$.

\begin{figure}[t]
    \centering
    \includegraphics[width=\linewidth]{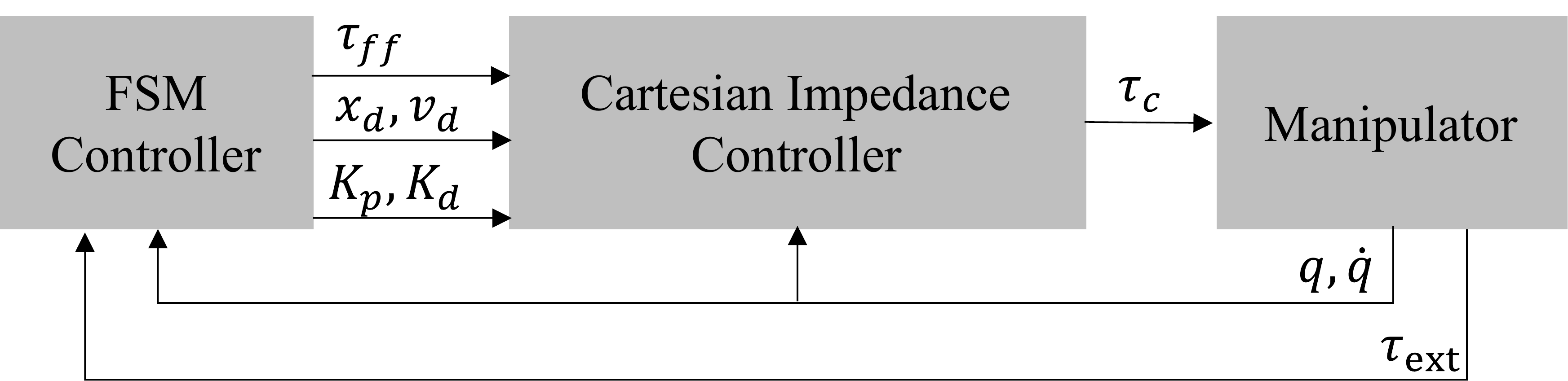}
    \caption{\textbf{An overview of the control framework.} The FSM controller sends high level command to the Cartesian Impedance controller, which then sends the commanded torque to the manipulator.}
    \label{fig:system_diagram}
    \vspace{-15px}
\end{figure}

The overall control framework is shown in Fig.~\ref{fig:system_diagram}.  The Finite-State-Machine controller sends high-level commands ($\feedforwardTau, x_d, v_d, K_p, K_d$) to the Cartesian Impedance controller, which computes the commanded torque $\commandTorque$ on the robot. The gains $K_p$ and $K_d$ are different at different FSM states.

\section{Simulation validation}
To validate the effectiveness of our proposed search strategy and the overall control framework, we run comprehensive simulation experiments. In particular, we compare our proposed search strategy with the spiral search method for 600 different initial displacements $\displacement$.  The simulation platform used is \texttt{Drake}\cite{drake}. To improve the authenticity of the contact simulation, we use the Hydroelastic contact \cite{Elandt2019IROS} to model the contact between the peg and the hole. 

\begin{figure}[h]
    \centering
    \begin{subfigure}[b]{0.250\textwidth}
        \includegraphics[width=\textwidth]{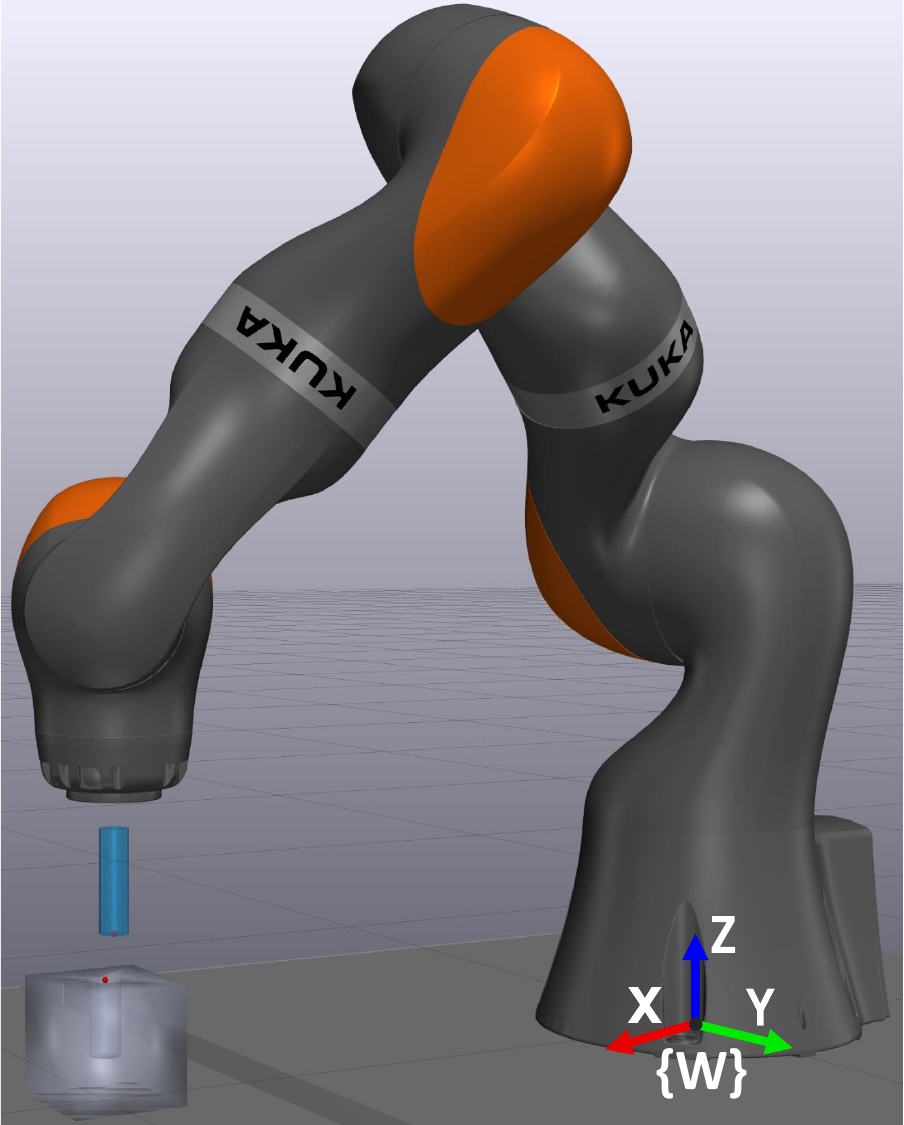}
        % \caption{}
        \label{}
    \end{subfigure}
    \hspace{0.001\textwidth}
    \begin{subfigure}[b]{0.16\textwidth}
        \includegraphics[width=\textwidth]{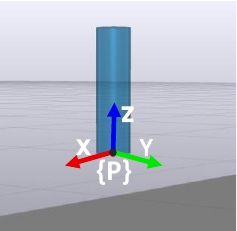}
        % \vphantom{\includegraphics[width=\textwidth]{figures/simulation_setup.pdf}}
        \includegraphics[width=\textwidth]{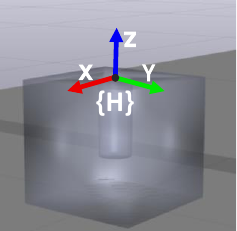}
        % \caption{}
        \label{}
    \end{subfigure}
    \vspace{-11px}
    \caption{\textbf{Setup of the simulation.} The world frame $\worldFrame$, the peg frame $\pegFrame$, and the hole frame $\holeFrame$ are shown.}
    
    \label{fig:simulation_setup}
    \vspace{-15px}
\end{figure}

\begin{figure}[b]
    \centering
    \includegraphics[width=0.8\linewidth]{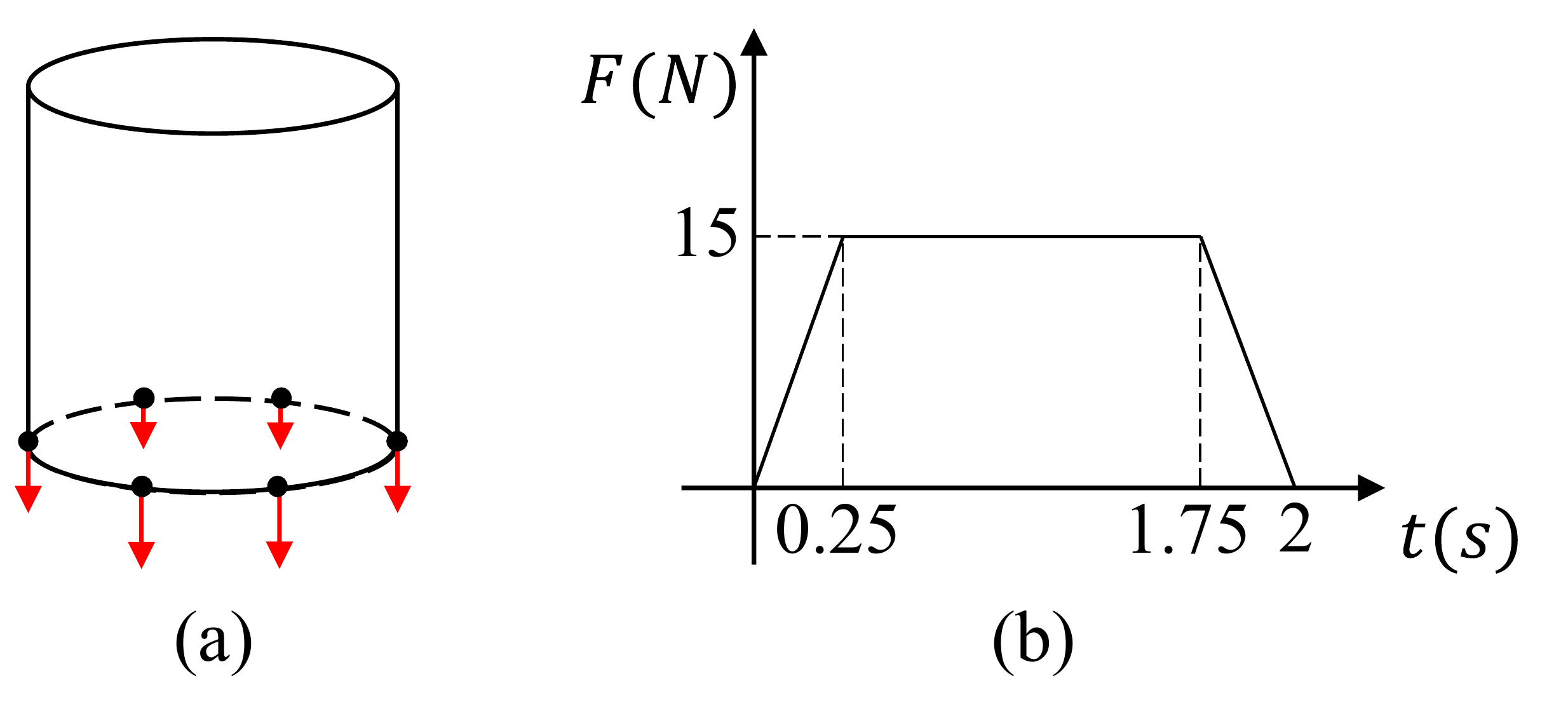}
    \vspace{-5px}
    \caption{\textbf{Illustration of action primitives $a_{f}^i$.} (a) Locations of the action primitives $a_f^i, i=1,\dots,6$ in our implementation. (b) Force magnitude of the action primitive $a_f^i$.}
    
    \label{fig:SixForce}
    % \vspace{-15px}
\end{figure}
\subsection{Experiment Setup}
The simulation setup is shown in Fig.~\ref{fig:simulation_setup}. We use the Kuka Iiwa 14 of 7 degrees of freedom as the robot model. A peg with radius $r=10\text{mm}$ is rigidly fixed to the robot's end-effector. The clearance of the hole is $1$mm. Both hole and peg are chamferless. The FSM controller runs at a frequency of $50$ Hz, and the Cartesian Impedance controller runs at a frequency of $1$ kHz. At the beginning of each experiment, the peg frame $\pegFrame$ is 3cm above the hole frame $\holeFrame$. We compare our proposed search strategy and the spiral search strategy for different initial displacements $\displacement = (\delta x, \delta y) = (\delta r \cos(\theta), \delta r \sin(\theta))$. In particular, $\delta r$ ranges uniformly from $6.67mm$ to $15mm$ for 10 values and $\theta$ ranges uniformly from $-\pi$ to $\pi$ uniformly for 60 values. Therefore, in total, 600 experiments are performed for each search strategy.

\subsection{Implementation Details}

In the implementation of our proposed search strategy, we choose the action primitives $a_f^i, i=1\dots K$ as downward forces distributed uniformly around the peg's edge, with $K=6$, as shown in Fig.~\ref{fig:SixForce}a. The force magnitude of $a_f^i$ is shown in Fig.~\ref{fig:SixForce}b. In the \textit{Move} state, the robot moves the peg directly towards the estimated hole's position in a straight line while applying a downward force of 5N at $\pegFrame$. No additional overlaid oscillation is added during the \textit{Move} state. In the comparison experiments, the \textit{Tilt} and \textit{Move} states of the FSM controller are replaced with a \textit{Spiral} state. In the \textit{Spiral} state, the robot moves the peg following a spiral path on the hole's surface while applying a downward force of 5N at $\pegFrame$. For all experiments, we do not reset the robot after the \textit{Fail} state is reached. The robot can only attempt to insert once in each experiment.

\begin{table*}[t]
    \tabcolsep = 2.5pt
    \centering
    \captionsetup{justification=centering}
    \caption{\textbf{Comparison between the proposed method and the spiral search method on different metrics.} \\ The Statistics for each method's 600 experiment results are provided.}
    \vspace{-1mm}
    \begin{tabular}{l|c|c|c|c|c}
    \toprule
% \hline
    \diagbox[width=7em]{methods}{metric} & $\rho$ & $T_s$(\text{mean}$\pm$\text{std}) & $l_s$(\text{mean}$\pm$\text{std}) & $T_c$(\text{mean}$\pm$\text{std}) & $\epsilon$(\text{mean}$\pm$\text{std})\\ 
    \midrule
    \midrule

    % \vspace{5px}
    \textbf{Our method }   & $596/600$ & $10.17\pm 3.24s$   & $1.33\pm 0.28cm$   & $12.52\pm 3.36s$   & $1.38\pm 1.24^{\circ}$ \\ 
    
    Spiral search & $509/600$ & $102.88\pm 20.46s$ & $46.84\pm 13.14cm$ & $106.64\pm 20.46s$ & \textbf{--} \\  
    \bottomrule
    \end{tabular}
    \label{tab:statistics}
    % \vspace{-15px}
\end{table*}

\begin{figure*}[bt]
    \begin{subfigure}[b]{.18\textwidth}
    \centering
    \includegraphics[width=\textwidth]{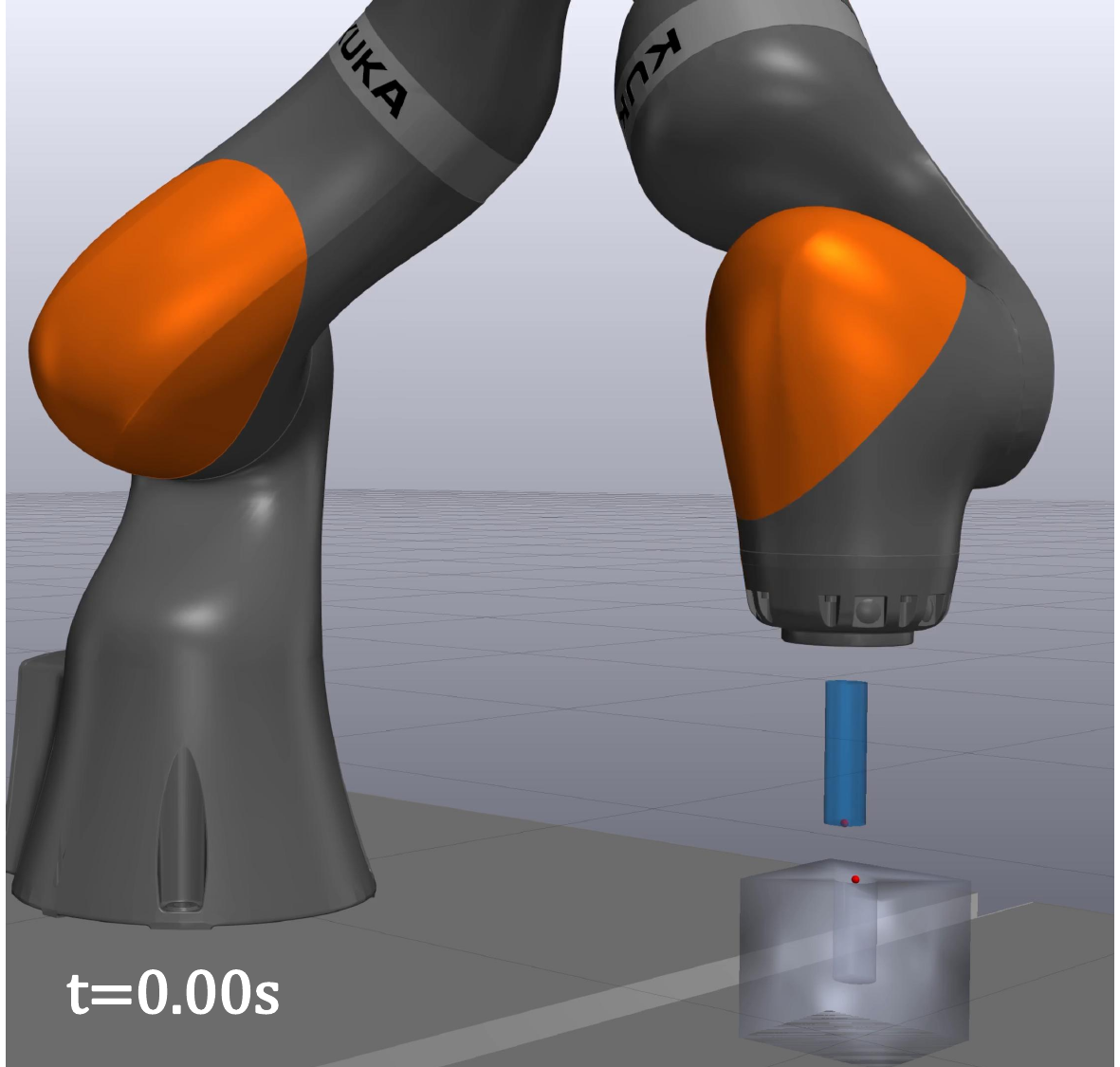}
    \end{subfigure}
    \hfill
    \begin{subfigure}[b]{.18\textwidth}
    \centering
    \includegraphics[width=\textwidth]{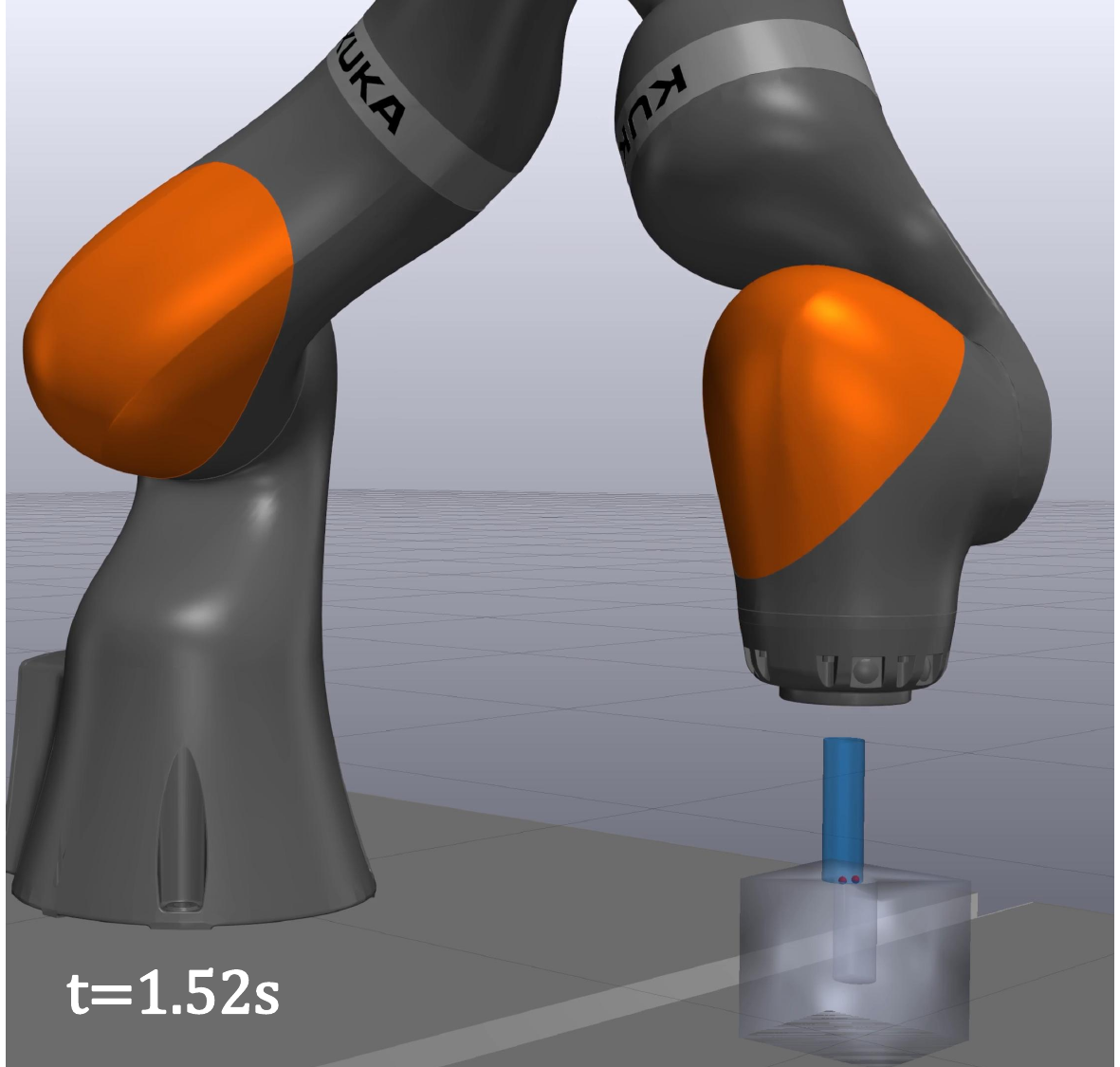}
    \end{subfigure}
    \hfill
    \begin{subfigure}[b]{.18\textwidth}
    \centering
    \includegraphics[width=\textwidth]{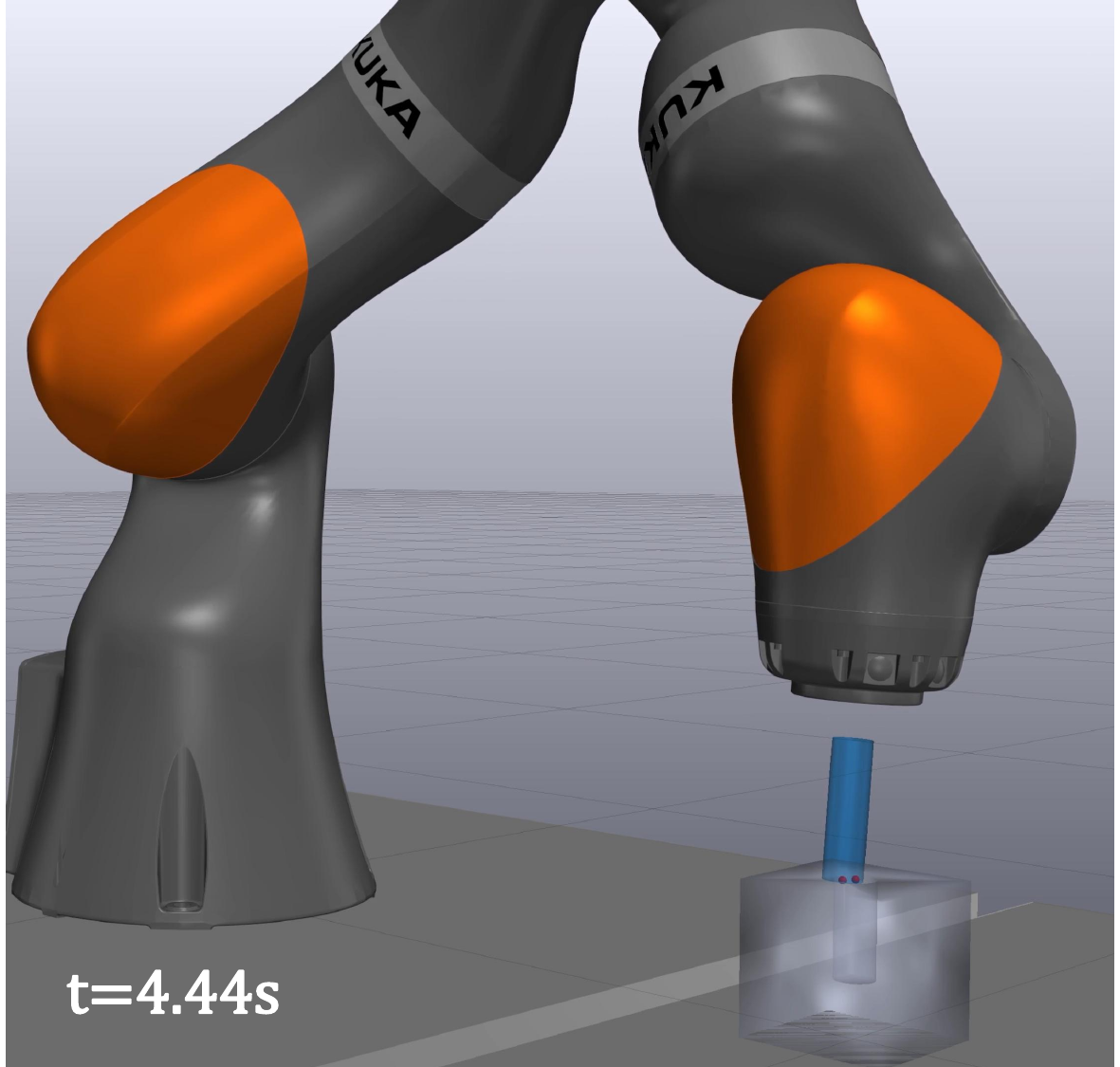}
    \end{subfigure}
    \hfill
    \begin{subfigure}[b]{.18\textwidth}
    \centering
    \includegraphics[width=\textwidth]{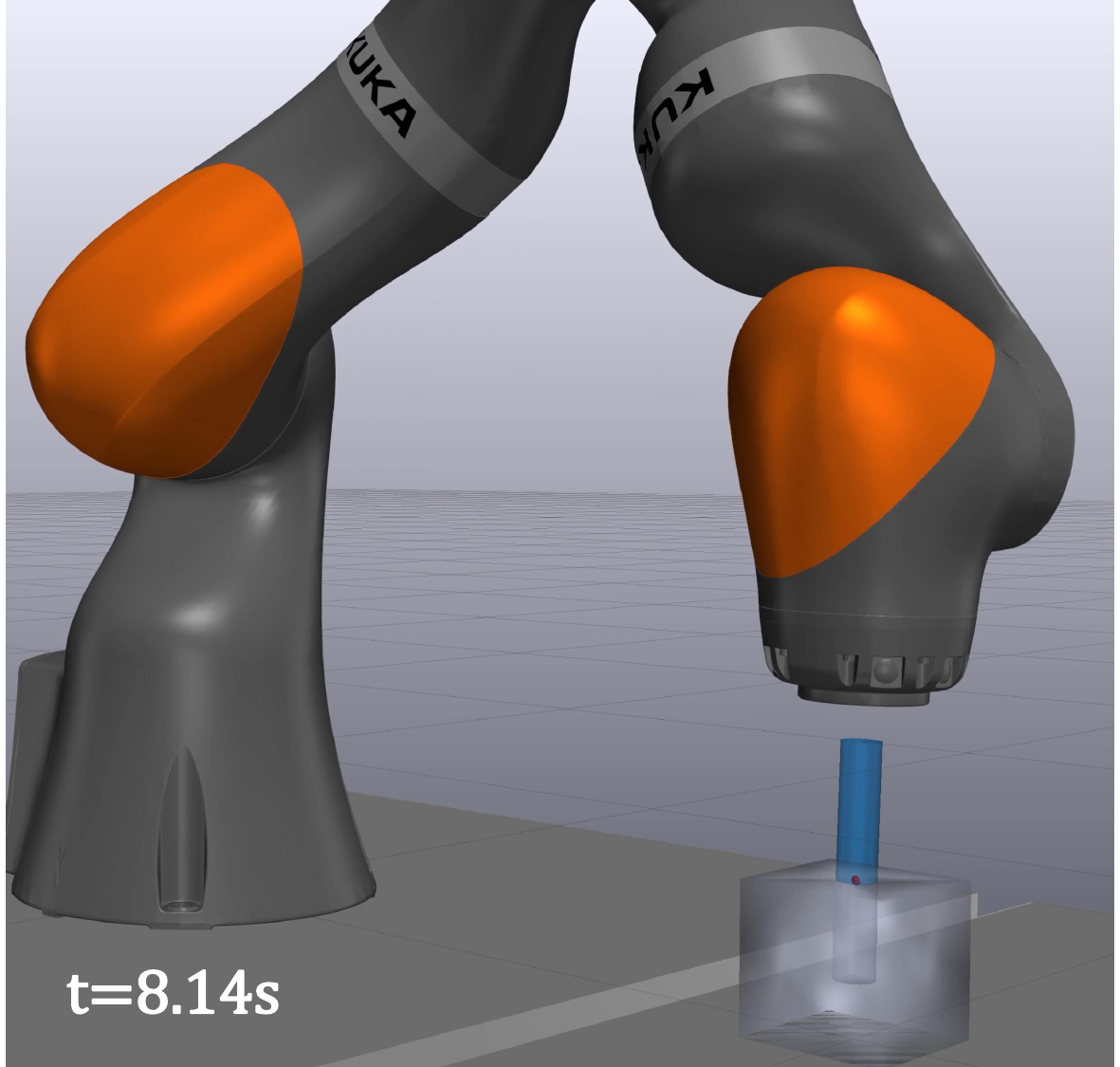}
    \end{subfigure}
    \hfill
    \begin{subfigure}[b]{.18\textwidth}
    \centering
    \includegraphics[width=\textwidth]{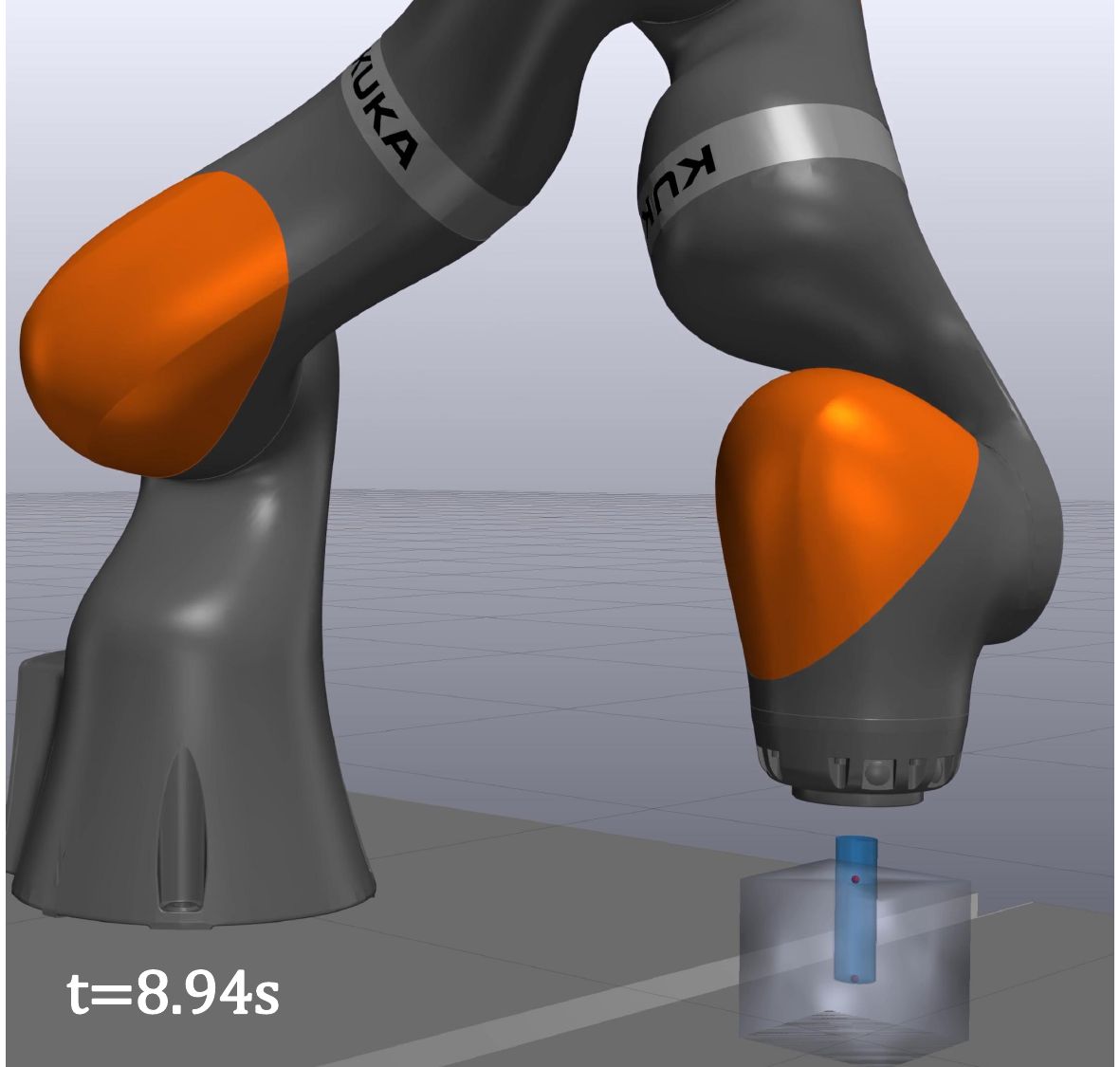}
    \end{subfigure}
    \hfill
    \caption{\textbf{Snapshots of one experiment with the proposed method.} Initial displacement is $(\delta r = 6.6mm, \theta=-137^\circ)$. As shown, the robot first lowers the peg to make contact with the hole. It then intentionally tilts the peg using action primitives $a_f^i$ to determine the hole's relative position. Once the relative position is found, the robot moves the peg straight toward the hole. Finally, when the peg is aligned with the hole, the robot fully inserts it into the hole.}
    \label{fig:experiment_snapshot}
    % \vspace{-10px}
\end{figure*}

\subsection{Metrics}
To evaluate the performance of the search strategies, we define 4 metrics: success rate, search time, search trajectory length, and cycle time. In addition, we define a metric, estimation accuracy, to evaluate the proposed search strategy's precision in determining the hole's position.
\subsubsection{Success Rate $\rho$}
One experiment is regarded as successful if the peg can be fully inserted into the hole. It is regarded as failed if the \textit{fail} state is reached or the maximum time is reached. The success rate of a search strategy is $\rho = N_s / N$, where $N_s$ is the number of successful experiments and $N = 600$ is the total number of experiments.
\subsubsection{Search Time $T_s$}
Search time $T_s = t_i - t_s$ is the total time of the search phase where $t_i$ is the time that \textit{Insert} state is entered and $t_s$ is the time that \textit{Tilt} or \textit{Sprial} state is entered. 
\subsubsection{Search trajectory length $l_s$}
The length of $\pegFrame$'s trajectory during the search phase.
\subsubsection{Cycle Time $T_c$}
The total time of the insertion process.

\subsubsection{Estimation accuracy $\epsilon$}
Denote the estimated hole's displacement as $\hat{\displacement}$. The estimation accuracy is defined as 
\begin{equation}
    \epsilon = \arccos ( \frac{\displacement^T\hat{\displacement}}{\norm{\displacement}\norm{\hat{\displacement}}})
\end{equation}
The unit of $\epsilon$ is in degree.

\begin{figure}[htbp]
  \centering
  \begin{subfigure}[b]{0.235\textwidth}
    \centering
    \includegraphics[width=\textwidth]{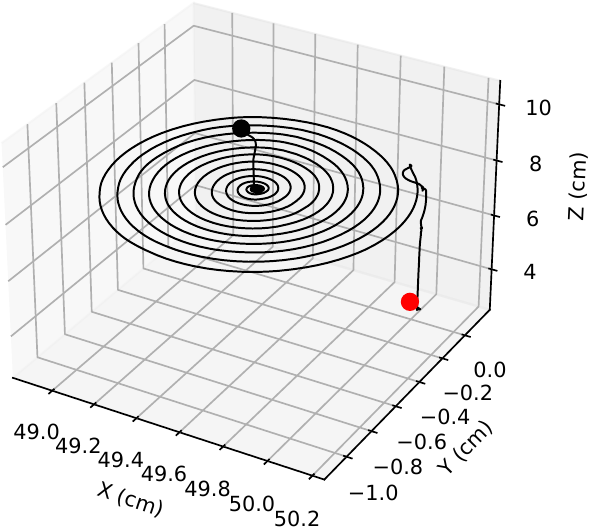}
    \caption{}
    \label{fig:spiral_trajectory}
  \end{subfigure}
  \hfill
  \begin{subfigure}[b]{0.235\textwidth}
    \centering
    \includegraphics[width=\textwidth]{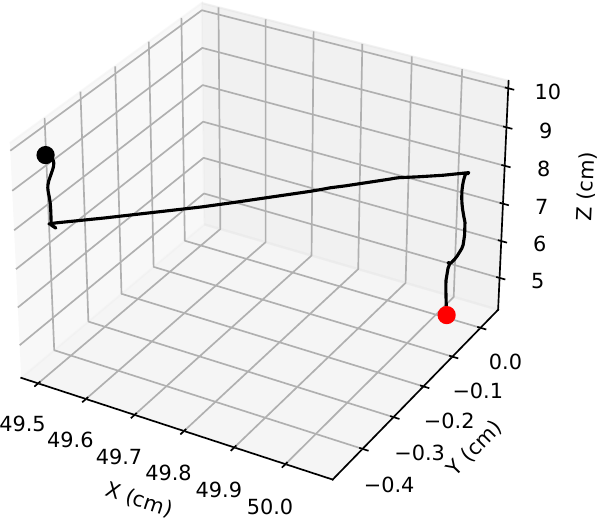}
    \caption{}
  \label{fig:our_trajectory}
  \end{subfigure}
  \caption{\textbf{Trajectories of the peg frame $\pegFrame$ using the spiral search method and our method.} The initial displacement for both methods is initial displacement $(\delta r = 6.6mm, \theta=-137^\circ)$. The black dot is the initial position. The red dot is the goal position. (a)Spiral search's trajectory. (b)Our method's trajectory.}
  \vspace{-15px}
\end{figure}

\subsection{Results}
The experiment results are summarized in Table.~\ref{tab:statistics}. Our method achieved 596 successful experiments out of 600, while the spiral search method achieved 509 successful experiments out of 600. The average cycle time for our method is 12.5s, while for the spiral search method is 106.6s. The average search time for our method is 10.2s, while for the spiral search method is 102.9s. The spiral search method needs to take a much longer time to locate the hole compared with our proposed method. We observe that the search phase accounts for most of the time for both our method and the spiral search method. Moreover, our method's average search trajectory length is much shorter than the spiral search method: 1.3cm versus 46.8cm. Snapshots of one experiment using our proposed method are shown in Fig.~\ref{fig:experiment_snapshot}. The trajectory of the peg frame $\pegFrame$ during this experiment is shown in Fig.~\ref{fig:our_trajectory}. The trajectory of the peg frame $\pegFrame$ using spiral search from the same initial position is shown in Fig.~\ref{fig:spiral_trajectory}. With our method, the peg follows a much more efficient trajectory. Our method accurately estimates the hole's position, with average $\epsilon = 1.38^{\circ}$.
% The average estimation accuracy $\epsilon$ of our method is $1.38^{\circ}$.

\subsection{Discussion}
The results show that our proposed method can accurately estimate the hole's position and insert the peg into the hole with a high success rate. Compared with the spiral search method, our proposed method achieves a higher success rate and is much more efficient in search time and trajectory length. In our implementation, during the \textit{Move} state, the peg follows a straight line towards the hole, imposing a strict estimation accuracy requirement. To improve the robustness of our method, small overlaid oscillation can be added to the peg's motion in the \textit{Move} state.

\section{Conclusion}\label{sec:con} 
In this work, we present a novel search strategy for the robotic insertion task. Our proposed search strategy is based on a systematical POMDP formulation of the search problem and the analysis of the contact configuration's static stability. The proposed search strategy is realized with a control framework combining a high-level Finite-State-Machine controller and a low-level Cartesian Impedance Controller. Our method is suitable for torque-controlled robots and only relies on the robot's proprioceptive sensing ability. Extensive comparison experiments with the baseline method are conducted to validate the effectiveness of our method. The results show that the proposed method is robust to different initial displacement errors and exhibits a high success rate. In addition, our method is more efficient and has both a shorter search time and search trajectory length.

In the future, we plan to test the proposed method in real hardware. Sensor noise and joint friction within a real-world robotic system pose potential challenges to our proposed method. Extending our method to consider complex-shape peg and orientation error is also a potential future direction.
\bibliographystyle{ieeetr}
\bibliography{mybibfile}

\end{document}

%% file: packages.tex
\usepackage{amsthm}
\usepackage{times,multirow,caption,float}
\let\labelindent\relax
\usepackage{enumitem}
\usepackage{microtype}
\usepackage{wrapfig}
\usepackage{graphicx}
\usepackage{epsfig,xspace,layout}
\usepackage[ruled, vlined]{algorithm2e}
\usepackage{svg}
\usepackage{color}
\usepackage{amsfonts}
\usepackage{times}
\usepackage{amssymb}
\usepackage{amsmath}
\usepackage{bm}
\usepackage{rotating}
\usepackage{epsfig}
\usepackage{mathrsfs}
\usepackage{caption}
\usepackage{subcaption}
\usepackage{mathtools}
\usepackage{makeidx}
\usepackage{multirow} 
\usepackage{dblfloatfix}
\usepackage{threeparttable}
\usepackage{dsfont}
\usepackage[font={small}]{caption}
\usepackage{cleveref}
\usepackage{tikz}
\usetikzlibrary{shapes,arrows,positioning,calc}
\usepackage{siunitx}
\usepackage{url}
\usepackage{cite}

\usepackage{diagbox}
\usepackage{tabularx}
\usepackage{booktabs}

%% file: newcommands.tex
\theoremstyle{definition}

\newtheorem{problem}{Problem}
\theoremstyle{remark}
\newtheorem{remark}{Remark}

\DeclareMathAlphabet{\mathpzc}{OT1}{pzc}{m}{it}

\DeclareFontFamily{U}{jkpmia}{}
\DeclareFontShape{U}{jkpmia}{m}{it}{<->s*jkpmia}{}
\DeclareFontShape{U}{jkpmia}{bx}{it}{<->s*jkpbmia}{}
\DeclareMathAlphabet{\mathfrak}{U}{jkpmia}{m}{it}

\newcommand{\norm}[1]{\left\lVert#1\right\rVert}

\newcommand{\q}{q}
\newcommand{\dq}{\dot{q}}
\newcommand{\ddq}{\ddot{q}}
\newcommand{\R}{\mathbb{R}}

\newcommand{\mass}{M}
\newcommand{\coriolis}{C}
\newcommand{\gravityTorque}{\tau_g}
\newcommand{\commandTorque}{\tau_c}
\newcommand{\jacobian}{J}
\newcommand{\wrench}{F}

\newcommand{\worldFrame}{\{W\}}
\newcommand{\pegFrame}{\{P\}}
\newcommand{\holeFrame}{\{H\}}
\newcommand{\eg}{\textit{e.g.}}